\documentclass[11pt,a4paper]{article}
  \usepackage[hyperref]{naaclhlt2018}
  \usepackage{times}
  \usepackage{latexsym}
  \usepackage{url}
  \usepackage{amsmath}
  \usepackage{amstext}
  \usepackage{amssymb}
  \usepackage{graphicx}
  \usepackage{tabularx}
  \usepackage{color}
  \usepackage{subcaption}
  \usepackage{enumitem}
  \usepackage{bm}
  \usepackage{multirow}
  \usepackage{paralist}

  \aclfinalcopy % Uncomment this line for the final submission
  \def\aclpaperid{1776} %  Enter the acl Paper ID here

  \newcommand{\bMM}{{\bf M}}
  \newcommand{\bTT}{{\bf T}}
  \newcommand{\bWW}{{\bf W}}
  \newcommand{\bB}{{\bf b}}
  \newcommand{\bE}{{\bf e}}
  \newcommand{\bP}{{\bf p}}
  \newcommand{\bR}{{\bf r}}
  \newcommand{\bY}{{\bf y}}
  \newcommand{\bZ}{{\bf z}}
  
  \renewenvironment{enumerate}[1]{\begin{compactenum}#1}{\end{compactenum}}
  \title{Using Aspect Extraction Approaches \\to Generate Review Summaries and User Profiles}

  \author{Christopher Mitcheltree\thanks{\enskip Equal contribution.}\quad Skyler Wharton\footnotemark[1]\quad Avneesh Saluja \\
    Airbnb AI Lab\\
    San Francisco, CA, USA \\
    {\tt firstname.lastname@airbnb.com}
  }
  \date{}

\begin{document}
  \maketitle
  \begin{abstract}
  Reviews of products or services on Internet marketplace websites contain a rich amount of information.
  Users often wish to survey reviews or review snippets from the perspective of a certain aspect, which has resulted in a large body of work on aspect identification and extraction from such corpora.
  In this work, we evaluate a newly-proposed neural model for aspect extraction on two practical tasks.
  The first is to extract canonical sentences of various aspects from reviews, and is judged by human evaluators against alternatives.
  A $k$-means baseline does remarkably well in this setting.
  The second experiment focuses on the suitability of the recovered aspect distributions to represent users by the reviews they have written.
  Through a set of review reranking experiments, we find that aspect-based profiles can largely capture notions of user preferences, by showing that divergent users generate markedly different review rankings.
  \end{abstract}

  \section{Introduction}
  \label{sec:intro}

  Aspect extraction has traditionally been associated with the sentiment analysis community \cite{Liu2012, Pontiki2016}, with the goal being to decompose a small document of text (e.g., a review) into multiple facets, each of which may possess their own sentiment marker.
  For example, a restaurant review may comment on the ambiance, service, and food, preventing the assignment of a uniform sentiment over the entire review.
  A common approach to aspect extraction is to treat the aspects as latent variables and utilize latent Dirichlet allocation (LDA; \newcite{Blei2003}) to extract relevant aspects from a collection of documents in an unsupervised \cite{Titov2008,Brody2010} or semi-supervised \cite{Mukherjee2012} fashion.
  Subsequent research has taken the latent variable approach further by encoding more complicated dependencies between aspects and sentiment \cite{Zhao2010}, or between aspects, ratings, and sentiment \cite{Diao2014}, using probabilistic graphical models \cite{Koller2009} to jointly learn the parameters.

  However, it has been argued that the coherence of aspects extracted from the family of LDA-based approaches is low; words clustered together within a specific aspect are often unrelated, which can be attributed to the lack of word co-occurrence information in these models \cite{Mimno2011}, since conventional LDA assumes each word in a document is generated independently.
  Recently, \newcite{He2017} proposed a neural attention-based aspect extraction (ABAE) approach, which like LDA, is an unsupervised model.
  The starting point is a set of word embeddings, where the vector representation of the word encapsulates co-occurrence\footnote{words that co-occur with each other get mapped to points close to each other in the embedding space \cite{Harris1968,Schutze1998}.}.
  The embeddings are used to represent a sentence as a bag-of-words, weighted with a self-attention mechanism \cite{Lin2017}, and learning amounts to encoding the resulting attention-based sentence embedding as a linear combination of aspect embeddings, optimized using an autoencoder formulation (\S\ref{sec:bg}).
  The attention mechanism thus learns to highlight words that will be pertinent for aspect identification.

  In this work, we apply the ABAE model to a large corpus of reviews on Airbnb\footnote{\texttt{www.airbnb.com}}, an online marketplace for travel; users (guests) utilize the site to find accommodation (listings) all around the world, and a large number of these guests write reviews of the listing post-stay.
  We first provide additional details on the workings of the ABAE model (\S\ref{sec:bg}).
  ABAE is then applied to two tasks: the first (\S\ref{sec:tasks_summarize}) is to extract a representative sentence from a set of listing-specific reviews for a number of pre-defined aspects e.g., cleanliness and location, with the efficacy of extractive summarization evaluated by humans (\S\ref{sec:eval_icon}).
  Surprisingly, we find that the $k$-means baseline performs very well on aspects that occur more frequently, but ABAE may be better for infrequent aspects.

  In the second task (\S\ref{sec:tasks_profile}), we analyze the suitability of aspect embeddings to represent guest profiles.
  The hypothesis is that the content of guest reviews reveals the guest's preferences and priorities \cite{Chen2015}, and that these preferences correspond to extracted aspects.
  We investigate several ways to aggregate sentence-level aspect embeddings at the review and user levels and compute distances between user aspect and listing review embeddings, in order to personalize listing reviews by reranking them for each user.
  The correlation between guest profile distances (computed on pairs of guests) and review rank distances (computed on pairs of ordinal rankings over reviews) is then measured to evaluate our hypothesis (\S\ref{sec:eval_ranking}).
  We find a robust relationship between distances in the two spaces, with the correlation increasing at finer granularities like sentences compared to reviews or listings.

  \section{Background}
  \label{sec:bg}

  To start, we provide a brief background of the ABAE model.
  For additional details, please refer to the original paper \cite{He2017}.
  At a high level, the ABAE model is an autoencoder that minimizes the reconstruction error between a weighted bag-of-words (BoW) representation of a sentence (where the weights are determined by a self-attention mechanism) and a linear combination of aspect embeddings.
  The linear combination represents the probabilities of the sentence belonging to each of the aspects.

  The first step in ABAE is to compute the embedding $\bZ_s \in \mathbb{R}^d$ for a sentence $s$:
  \begin{align*}
    \bZ_s &= \sum_{i=1}^n a_i \bE_{w_i}
  \end{align*}
  where $\bE_{w_i}$ is the word embedding $\bE \in \mathbb{R}^d$ for word $w_i$.
  As in the original paper, we use word vectors trained using the skip-gram model with negative sampling \cite{Mikolov2013}.
  The attention weights $a_i$ are computed as a multiplicative self-attention model:
  \begin{align*}
    a_i &= \textrm{softmax}(\bE_{w_i}^{\textrm{T}} \cdot \bMM \cdot \bY_s) \\
    \bY_s &= \sum_{i=1}^n \bE_{w_i}
  \end{align*}
  where $\bY_s$ is simply the uniformly-weighted BoW embedding of the sentence, and $\bMM \in \mathbb{R}^{d \times d}$ is a learned attention model.

  The next step is to compute the aspect-based sentence representation $\bR_s \in \mathbb{R}^d$ in terms of an aspect embeddings matrix $\bTT \in \mathbb{R}^{K \times d}$, where $K$ is the number of aspects:
  \begin{align*}
    \bP_s &= \textrm{softmax}(\bWW \cdot \bZ_s + \bB) \\
    \bR_s &= \bTT^{\textrm{T}} \cdot \bP_s
  \end{align*}
  where $\bP_s \in \mathbb{R}^K$ is the weight (probability) vector over $K$ aspect embeddings, and $\bWW \in \mathbb{R}^{K \times d}, \bB \in \mathbb{R}^K$ are parameters of a multiclass logistic regression model.

  The model is trained to minimize reconstruction error (using the cosine distance between $\bR_s$ and $\bZ_s$) with a contrastive max-margin objective function \cite{Weston2011}.
  In addition, an orthogonality penalty term is added to the objective, which encourages the aspect embedding matrix $\bTT$ to produce diverse (orthogonal) aspect embeddings.

  \section{Tasks}
  \label{sec:tasks}

  To evaluate the utility of ABAE, we craft two methods of evaluation that mimic the practical ways in which aspect extraction can be used on a marketplace website with reviews.

  \subsection{Extractive Summarization}
  \label{sec:tasks_summarize}

  The first task is a direct evaluation of the quality of the recovered aspects: we use ABAE to select review sentences of a listing that are representative of a set of preselected aspects, namely cleanliness, communication, and location.
  ``Cleanliness" refers to how clean the listing is, ``communication" refers to communication between the listing host and the guest, and ``location" refers to the qualities of or amenities in the listing's neighborhood.
  Refer to Table \ref{tab:aspect_words} for representative words for each aspect.
  Thus, aspect extraction is used to summarize listing reviews along several manually-defined topics.

  We benchmark the ABAE model's extracted aspects against those from two baselines: LDA and $k$-means.
  For each experimental setup, the authors assigned one of four interpretable labels (corresponding to the identified aspects and the ``other'' category) to each unlabeled aspect by evaluating the 50 words most associated with that aspect\footnote{Inter-annotator agreements for each setup are provided in Table \ref{tab:aspect_words}.}.
  LDA's topics are represented as distributions over words, so the most associated words correspond to those that occur with highest probability.
  For $k$-means and ABAE, each aspect is represented as a point in word embedding space\footnote{The aspect embeddings in ABAE are initialized using the $k$-means centroids.}, so we retrieve the 50 closest words to each point using cosine distance as a measure.

  After aspect identification, we infer aspect distributions for the review sentences of an unseen test set of listings.
  For LDA, identification simply amounts to computing the (approximate) posterior over topic mixtures for a set of review sentences, and selecting the sentences with the highest probability in the specified aspect.
  For $k$-means, each sentence is represented as a uniformly-weighted BoW embedding, and we retrieve the sentences that are closest to the centroids that correspond to our preselected aspects.
  ABAE is similar, except the self-attention mechanism is applied to compute an attention-based BoW embedding, and we retrieve the sentences closest to the aspect embeddings corresponding to our aspects of interest.
  For some aspects (e.g., location and communication), there is a many-to-one mapping between the recovered word clusters and the aspect label.
  In these cases, we compute the average of the aspect embeddings, and the closest sentences to the resulting points are retrieved.

  In our selection process, we retrieve the three most representative sentences (across all reviews of that listing) for each aspect.
  Three human annotators then evaluated the appropriateness of the selected aspect for each sentence via binary judgments.
  For example, an evaluator was presented with the sentence ``Easy to get there from center of city by subway and bus.'', along with the inferred aspect (\emph{location}), for which a binary ``yes/no'' response suffices.
  Results aggregated by experimental setup and aspect are presented in \S\ref{sec:eval_icon}.

  \subsection{Aspects as Profiles}
  \label{sec:tasks_profile}

  An aspect extraction model provides a distribution over aspects for each sentence, and we can consider these distributions as interpretable sentence embeddings (since we can assign a meaning corresponding to an aspect for each of the extracted word clusters).
  These embeddings can be used to provide \emph{guest profiles} by aggregating aspect distributions over review sentences that a user has written across different listings on the website.
  Such profiles arguably capture finer-grained information about guest preferences than an aggregate star rating across all aspects.
  Star ratings are also heavily positively-biased: more than 80\% of reviews on Airbnb rate the maximum of 5 stars.

  There are many conceivable ways to aggregate the sentence distributions, with some of the factors of variation being:
  \begin{enumerate}
      \item level of hierarchy: is the guest considered to be a bag-of-reviews (BoR), sentences (BoS), or words i.e., do we weight longer sentences or reviews with more sentences higher when computing the aggregated representation?
      \item time decay: how do we treat more recently written reviews compared to earlier ones?
      \item average or maximum: is the aggregate representation a (weighted) average, or should we consider the maximum value across each aspect and renormalize?
  \end{enumerate}
  The same considerations also arise when computing the representations of the objects to be ranked e.g., a listing embedding as the aggregation of its component sentence embeddings.

  For our evaluation (\S\ref{sec:eval_ranking}), guest profiles are computed by averaging distributions across a guest's review sentences uniformly (BoS), equivalent to a BoR representation weighted by the review length (number of sentences).
  We also experimented with a BoR representation using uniformly-weighted reviews, and the results are very similar to the BoS representation.
  We considered computing the guest profile by utilizing the maximum value (probability) of each aspect dimension across all review sentences written by the user and renormalizing the resulting embedding using the softmax function, but this approach resulted in high-entropic guest profiles with limited use downstream.
  More complex aggregation functions, like using an exponential moving average to upweight recent reviews, is an interesting future direction to explore.

  \section{Evaluation}
  \label{sec:eval}

  We now look at the qualitative and quantitative performance of ABAE across the two tasks.
  After providing statistics on the review corpus that forms the basis of our evaluation, we qualitatively analyze the recovered aspects of the model, compared to $k$-means and LDA baselines.
  On a heldout evaluation set, human evaluators assessed whether the model-extracted aspects correspond to their understanding of the predefined ones by inspecting the top-ranked sentences for each aspect.
  Furthermore, the quality of the guest profile embeddings was evaluated by looking at the correlation between distances in the aspect space and the ordinal position of reviews on a given listing page, with the hypothesis that guests who write reviews with divergent content or aspects should receive rankings that are very different.

  Our experiments were implemented using the pyTorch package\footnote{\texttt{http://pytorch.org/}}.
  Word vectors were trained using Gensim \cite{Rehurek2010} with 5 negative samples, window size 5, and dimension 200, and Scikit-learn \cite{Pedregosa2011} was used to run the $k$-means algorithm and LDA with the default settings.
  For ABAE, we used Adam with a learning rate of 0.001 (and the default $\beta$ parameters) with a batch size of 50, 20 negative samples, and an orthogonality penalty weight of 0.1.
  All experiments were run on an Amazon AWS \texttt{p2.8xlarge} instance.

  \subsection{Datasets}
  \label{sec:eval_data}

  The corpus was extracted from all reviews across all listings on Airbnb written between January 1, 2010 and January 1, 2017.
  We used spaCy\footnote{\texttt{http://spacy.io/}} to segment reviews into sentences and remove non-English sentences.
  All sentences were subsequently preprocessed in the same manner as \newcite{He2017}, which entailed restricting the vocabulary to the 9,000 most frequent words in the corpus after stopword and punctuation removal.
  From the resulting set, we randomly sampled 10 million sentences across 5.8 million guests and 1.8 million listings to form a training set, and used the remaining unsampled sentences to select validation and test sets for the human evaluation (\S\ref{sec:eval_icon}) and ranking (\S\ref{sec:eval_ranking}) experiments.

  To select datasets for human evaluation, we identified all listings with at least 50 and at most 100 reviews in all languages and filtered out any \emph{listing} in the training set, resulting in 900 listings which were split into validation and test sets.
  The validation set is used to select an appropriate number of aspects, by computing coherence scores \cite{Mimno2011} as the number of aspects is varied in the ABAE model (\S\ref{sec:eval_aspects}).
  The test set was used to extract review sentences that were presented to our human evaluators; we ensured that every listing in the test set has at least 3 non-empty English review sentences.

  For the ranking correlation experiments, we first identified users who had written at least 10 review sentences in our corpus and removed those users that featured in the training set from this list.
  We then selected 20 users uniformly at random to form our validation set i.e., to compute guest profiles for\footnote{The most prolific guest in this set had written 66 review sentences.}.
  A subset of the human evaluation test set was used to compute the correlation between aspect space and ranking order distances; we selected all listings that had at least 20 review sentences, resulting in 69 listings for evaluation.
  Table \ref{tab:corpus_stats} presents a summary of corpus statistics for all of the datasets used in this work.

  \begin{table}[h!]
    \small
      \begin{center}
        \begin{tabular}{p{0.1\linewidth}|c|rrrrr}
          \hline
          Set & Task & Tokens & Sentences & Guests & Listings \\
          \hline
          Train & - & 68.0M & 10.0M & 5.8M  & 1.8M \\
          Val &  \S\ref{sec:tasks_summarize} & 91,124 & 14,173 & 3719 & 721 \\
          Test &  \S\ref{sec:tasks_summarize} & 21,069 & 3389 & 920 & 168 \\
          Val &  \S\ref{sec:tasks_profile} & 3189 & 543 & 20 & 202 \\
          Test &  \S\ref{sec:tasks_profile} & 13,925 & 2269 & 587 & 69
        \end{tabular}
      \end{center}
      \caption{\small Corpus statistics for the datasets that we use. All numbers are computed after preprocessing.}
      \label{tab:corpus_stats}
    \end{table}

  \subsection{Recovered Aspects}
  \label{sec:eval_aspects}
  Table \ref{tab:coherence_scores} presents coherence scores for the ABAE model as we varied the number of aspects.
  Similar to \newcite{He2017}, we considered a ``document'' to be a sentence, but treating reviews as documents or all reviews of a listing as a document revealed similar trends.
% * <skyler.wharton@airbnb.com> 2018-04-13T16:41:58.943Z:
% 
% > ``document'' to be a sentence,
% I believe these are actually the numbers for when we considered a document to be a listing
% 
% ^.
  The table shows that coherence score improvements taper off after 30 aspects, so we chose this aspect value for further experiments.
% * <skyler.wharton@airbnb.com> 2018-04-13T17:00:41.979Z:
% 
% >  taper off after 30 aspects
% The numbers in the chart don't support this - the coherence scores continue to get larger for 40 aspects
% 
% ^.
  \begin{table}[h!]
  \small
    \begin{center}
      \begin{tabular}{p{0.16\linewidth}|p{0.12\linewidth}p{0.12\linewidth}p{0.12\linewidth}|p{0.12\linewidth}}
        \hline
        Num. & \multicolumn{3}{c|}{\bf Num. Representative Words} & Sum \\
        Aspects & 10 & 30 & 50 &  \\
        \hline
        5 & -125 & -1106 & -2829 & -4060 \\
        10 & -148 & -1244 & -3017 & -4409 \\
        15 & -126 & -1069 & -2656 & -3851 \\
        30 & -101 & -760 & -1917 & -2778 \\
        40 & -84 & -701 & -1765 & -2550
      \end{tabular}
    \end{center}
    \caption{\small Coherence scores as a function of the number of aspects and the number of representative words used to compute the scores (higher is better). The summed values indicate significant improvement from 15 to 30 aspects. For details on computing coherence score, see \newcite{Mimno2011}.}
% * <skyler.wharton@airbnb.com> 2018-04-13T16:23:41.246Z:
% 
% Maybe note that a higher coherence score is better
% 
% ^.
    \label{tab:coherence_scores}
  \end{table}
  \begin{table*}[t!]
  \small
    \begin{center}
      \begin{tabular}{lp{0.25\linewidth}p{0.25\linewidth}p{0.25\linewidth}r}
        \hline
         & \multicolumn{3}{c}{\bf Aspects} &  \\
        Setup & Location & Cleanliness & Communication & Fleiss' $\kappa$ \\
        \hline
        $k$-means & union, music, minute, dozen, quarter, chain, zoo, buffet, nord, theater (3 clusters, 10.2\%) & master, conditioners, boiler, fabric, roll, smelling, dusty, shutter, dirty, installed (1 cluster, 3.6\%) & welcomed, sorted, proactive, fix, checkin, prior, replied, process, communicator, ahead (3 clusters, 9.9\%) & 0.58 \\
        LDA & restaurant, location, flat, walk, away, back, short, minute, bus, come (4 clusters, 16.2\%) & house, comfortable, clean, bed, beach, street, part, modern, appartment, cool (1 cluster, 3.5\%) & helpful, arrival, wonderful, coffee, loved, use, warm, communication, friendly, got (4 clusters, 14.5\%) & 0.46 \\
        ABAE & statue, tavern, woodsy, street, takeaway, woodland, cathedral, specialty, idyllic, attraction (6 clusters, 18.4\%) & clean, neat, pictured, immaculate, spotless, stylish, described, uncluttered, tidy, classy (1 cluster, 3.5\%) & dear, u, responsive, greeted, instruction, communicative, sent, contract, attentive, key (3 clusters, 10.1\%) & 0.46
      \end{tabular}
    \end{center}
    \caption{\small Representative words for each aspect of interest across experimental setups, along with the number of clusters mapped to that aspect in parentheses as well as the percentage of validation set sentences assigned to that cluster (the remaining sentences were assigned to ``Other").
    For the aspects with multiple clusters, we select a roughly equal number of words from each cluster.  Misspellings are deliberate.}
    \label{tab:aspect_words}
  \end{table*}

  Next, for each 30-aspect experimental setup, we identified the word clusters corresponding to the set of preselected aspects by labeling each revealed cluster with a value from the set $\{cleanliness, communication, location, other\}$.
  Note that the mapping from clusters to identified aspects is many-to-one (i.e., multiple clusters for the same aspect were identified for two of the three aspects, namely location and communication.)
  In fact, the number of clusters associated with each aspect is a proxy for the frequency with which these aspects occur in the corpus.
  To verify this claim, we computed aspect-based representations (\S\ref{sec:tasks_summarize}) for each sentence in the validation set used for comparing coherence scores, and utilized these representations to compute sentence similarities to each cluster, followed by a softmax in order to assign fractional counts i.e., a soft clustering approach.
  For each setup, Table \ref{tab:aspect_words} provides the top 10 words associated with each aspect, the number of clusters mapped to that aspect, and the number of validation sentences assigned to the aspect.
  The location and communication aspects are 3 to 6 times more prevalent than the cleanliness aspect.

  Qualitatively, the ABAE aspects are more coherent, especially in the cleanliness aspect, and do not include irrelevant words (often verbs) that are not indicative of any conceivable aspect, like ``got'', ``use'', or ``come''.
  $k$-means selects relevant words to indicate the aspect, but the aspects are relatively incoherent compared to ABAE.
  LDA has a difficult time identifying relevant words, indicating the importance of the attention mechanism in ABAE.
  Interestingly, we found that the inter-annotator agreement (Fleiss' $\kappa$) was slightly higher for the $k$-means baseline, but all scores are in the range of moderate agreement.

  \begin{table}[h!]
    \small
      \begin{center}
        \begin{tabular}{l|rrp{0.1\linewidth}}
          \hline
          & \multicolumn{3}{c}{\bf Aspects} \\
          Setup & Loc & Clean & Comm \\
          \hline
          $k$-means & {\bf 0.85/0.68} & 0.30/0.26 & {\bf 0.62/0.43} \\
          LDA & 0.16/0.17 & 0.09/0.10 & 0.11/0.13 \\
          ABAE & 0.45/0.46 & {\bf 0.45/0.32} & 0.41/0.35 \\
        \end{tabular}
      \end{center}
      \caption{\small Precision@1 and precision@3 for the extractive summarization task, as judged by our human evaluators.}
      \label{tab:human_eval_results}
    \end{table}

  \subsection{Extracting Prototypical Sentences}
  \label{sec:eval_icon}
  Table \ref{tab:human_eval_results} presents precision@1 and precision@3 results for each experimental setup-aspect pair, as evaluated by our human annotators.
  There are a total of 168 listings $\times$ 3 experimental setups $\times$ 3 aspects $\times$ 3 sentences per aspect $= 4536$ examples to evaluate; we set aside 795 examples to compute inter-annotator agreement, resulting in 2042 examples per annotator.
  Fleiss' $\kappa = 0.69$, which is quite high given the difficulty of the task\footnote{The \emph{communication} aspect (referring to host responsiveness and timeliness) is often easily confused with the friendliness of the host or staff.}.
  
  \begin{figure*}[!h]
  \small
    \begin{center}
    \begin{subfigure}{0.67\columnwidth}
      \centering
      Ranking BoS listings
      \includegraphics[width=1.0\columnwidth,keepaspectratio=true,trim={0 0 0cm 0},clip]{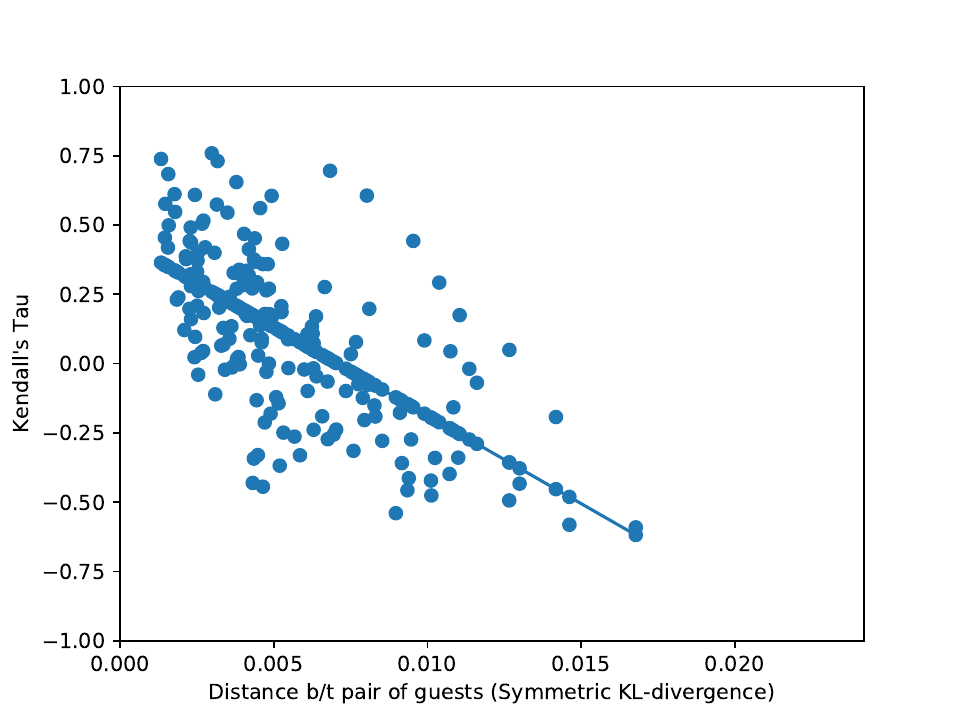}
      \caption{$R^2 = 0.39$.}
      \label{fig:guest_sent_listing_sent_abae}
    \end{subfigure}
    \begin{subfigure}{0.67\columnwidth}
      \centering
      Ranking BoS reviews \\(averaged over listings)
      \includegraphics[width=1.0\columnwidth,keepaspectratio=true,trim={0cm 0 0cm 0},clip]{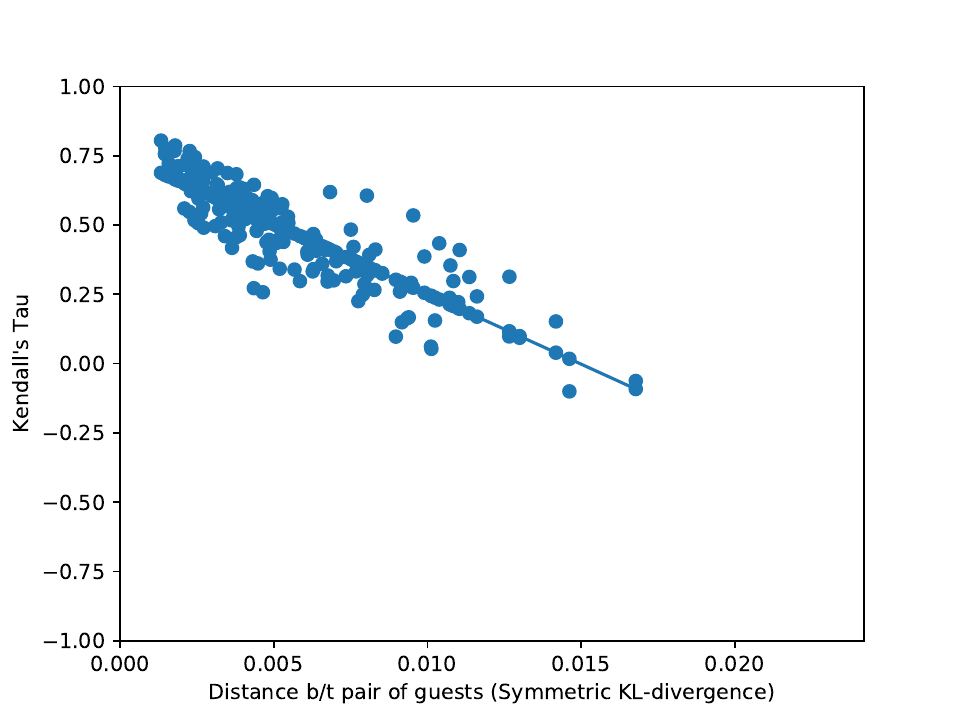}
      \caption{$R^2 = 0.73$.}
      \label{fig:guest_sent_review_listing_abae}
    \end{subfigure}
    \begin{subfigure}{0.67\columnwidth}
      \centering
      Ranking sentences \\(averaged over listings)
      \includegraphics[width=1.0\columnwidth,keepaspectratio=true,trim={0cm 0 0cm 0},clip]{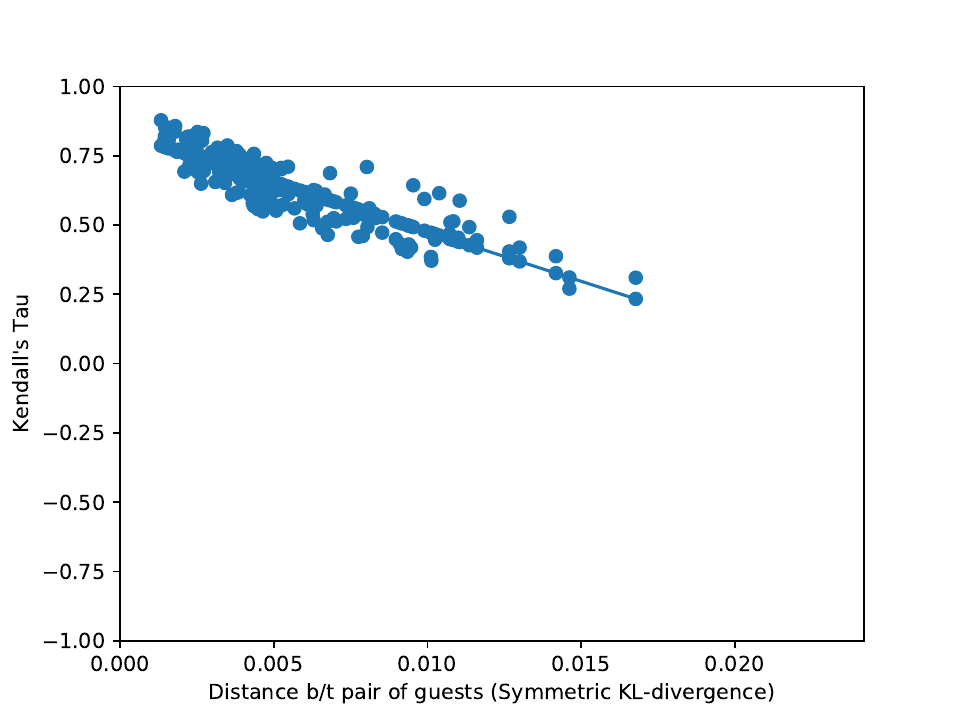}
      \caption{$R^2 = 0.75$.}
      \label{fig:guest_sent_sent_listing_abae}
    \end{subfigure}
    \end{center}
    \begin{center}
      \begin{subfigure}{0.67\columnwidth}
        \centering
        \includegraphics[width=1.0\columnwidth,keepaspectratio=true]{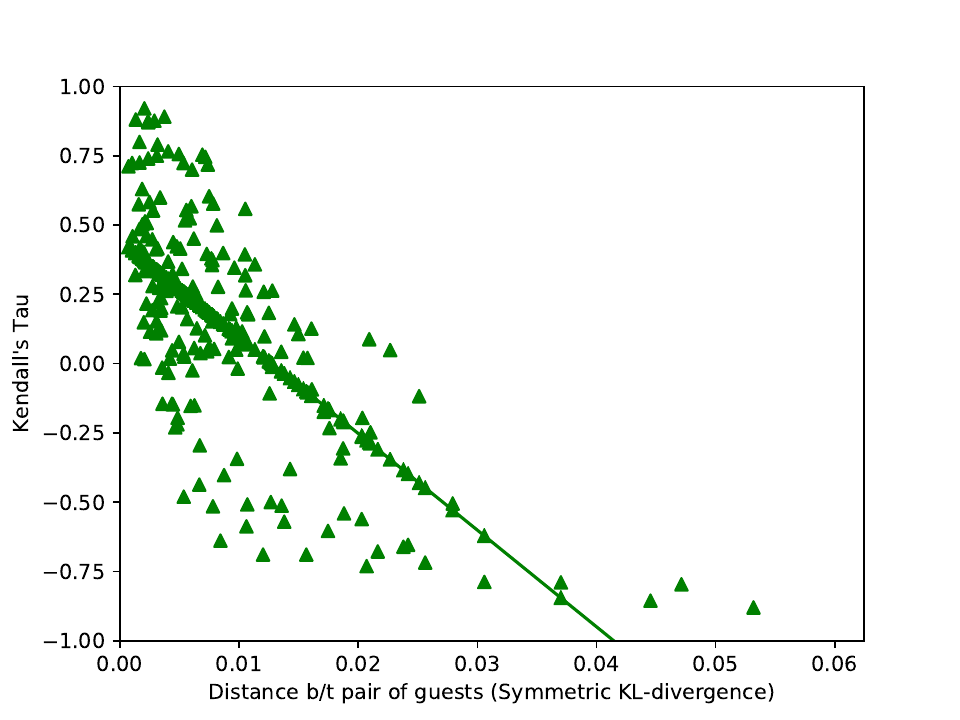}
        \caption{$R^2 = 0.46$.}
        \label{fig:guest_sent_listing_sent_kmeans}
      \end{subfigure}
      \begin{subfigure}{0.67\columnwidth}
        \centering
        \includegraphics[width=1.0\columnwidth,keepaspectratio=true]{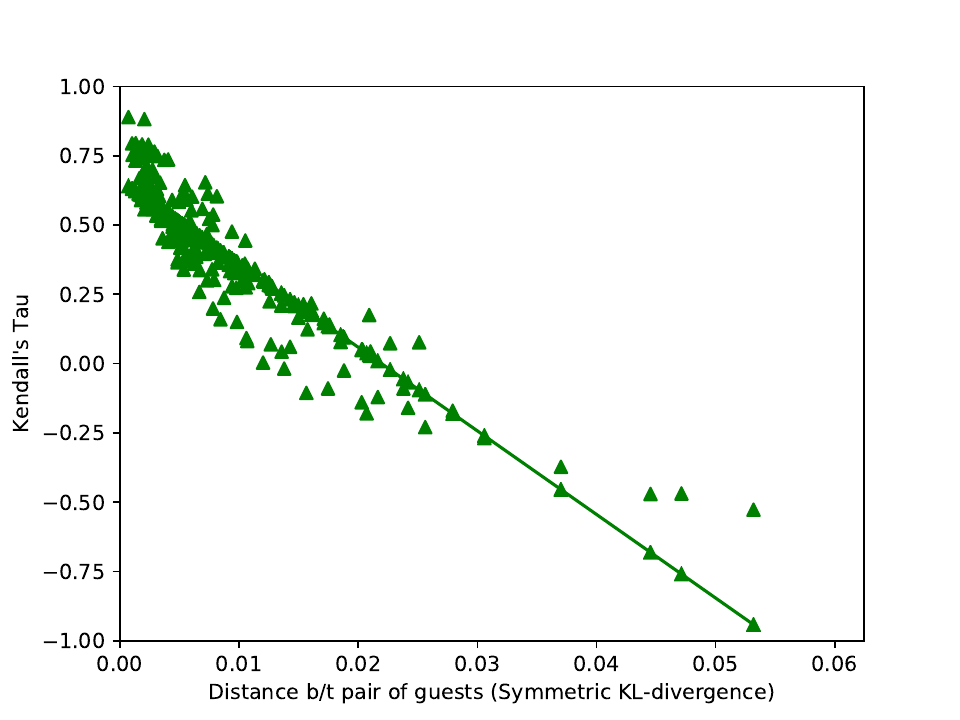}
        \caption{$R^2 = 0.81$.}
        \label{fig:guest_sent_review_listing_kmeans}
      \end{subfigure}
      \begin{subfigure}{0.67\columnwidth}
        \centering
        \includegraphics[width=1.0\columnwidth,keepaspectratio=true]{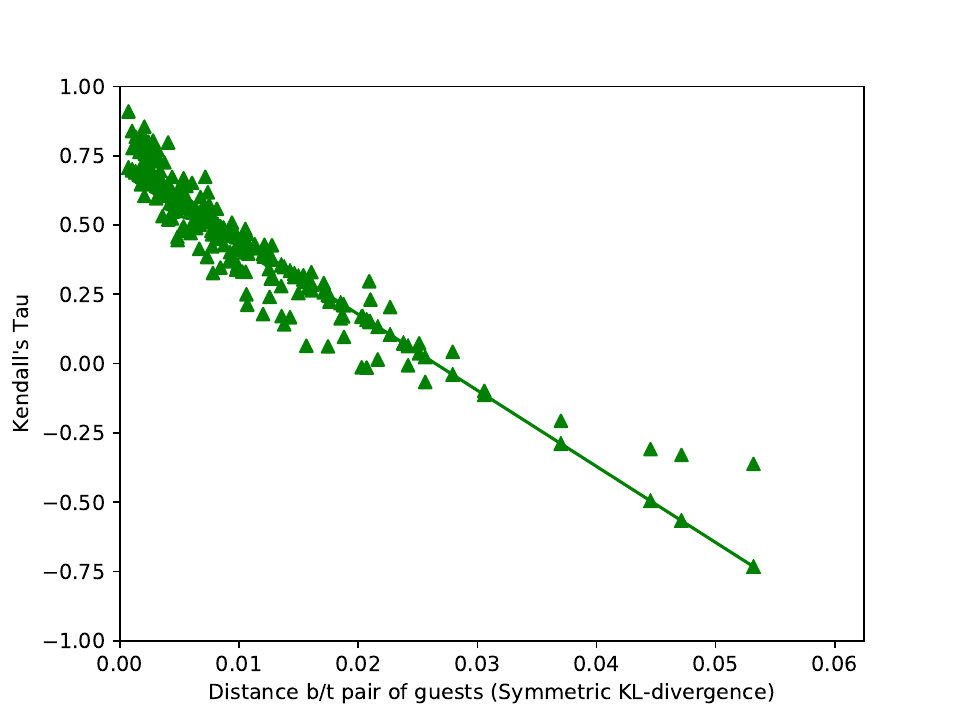}
        \caption{$R^2 = 0.84$.}
        \label{fig:guest_sent_sent_listing_kmeans}
      \end{subfigure}
    \end{center}
    \caption{\small Plots showing the relationship between distance in aspect space and ranking space, for guest profiles computed as bag-of-sentences representations. Figures \ref{fig:guest_sent_listing_sent_abae}, \ref{fig:guest_sent_review_listing_abae}, and \ref{fig:guest_sent_sent_listing_abae} are produced with the ABAE model, and figures \ref{fig:guest_sent_listing_sent_kmeans}, \ref{fig:guest_sent_review_listing_kmeans}, while \ref{fig:guest_sent_sent_listing_kmeans} are produced with the $k$-means model.}
    \label{fig:guest_sent_plots}
  \end{figure*}

  The most surprising result is that the $k$-means baseline is actually the strongest performer in the location and communication aspects.
  Nonetheless, the result is encouraging since it suggests that, for some aspects of interest to us, a simple $k$-means approach and uniformly-weighted BoW embeddings suffices.
  It is interesting to note that the strong baseline performance occurs with the aspects that occur more frequently in the corpus, as discussed in \S \ref{sec:eval_aspects}, suggesting that ABAE is more useful with aspects that occur more rarely in our corpus (e.g., cleanliness).
  For future work, we propose to evaluate this hypothesis in more depth by applying the approaches in this paper to the long tail of rarer aspects.
  The disappointing performance of LDA shows that its lack of awareness for word co-occurrence is damaging for aspect identification.

  \subsection{Review Ranking}
  \label{sec:eval_ranking}
  Figure \ref{fig:guest_sent_plots} presents results for the ranking correlation experiments with the ABAE and $k$-means models.
  The validation set is used to compute pairwise distances between all $\binom{20}{2} = 190$ guest pairs using the symmetric KL divergence, since guest profiles are probability distributions over aspects.
  This divergence forms the $x$-axis for our plots.
  We then rerank several objects of interest, and compute the rank correlation coefficient (Kendall's $\tau$) between pairs of rankings; this coefficient forms the $y$-axis for our plots.
  Lastly, the correlation between the distance in aspect space (between pairs of user profiles) and the distance in ranking space (between pairs of rankings over objects, as measured by Kendall's $\tau$) with $R^2$ values stated in the captions.

  With the guest profiles, we ranked the following objects using the symmetric KL divergence:
  \begin{enumerate}
    \item listings, where each listing is represented as a BoS (similar results were achieved when considering each listing as a BoR).
    \item reviews within a listing: for each guest pair and listing, we ranked the reviews using each guest's profile and computed Kendall's $\tau$ between the ranked pair of reviews.
    That score was then averaged over the 69 listings to yield a single score for each guest pair.
    \item sentences within a listing: similar to reviews within a listing, except Kendall's $\tau$ was computed over ranked sentences.
    The averaging step was the same as above.
  \end{enumerate}
  Since ABAE extracts aspects at the sentence-level, we would expect to see sentence-based representations result in higher correlations than other representations.
  Indeed, if we rank smaller units (i.e., sentences vs. listings), the correlation with distances in aspect space is higher (0.75 vs. 0.39 in the case of ABAE, 0.84 vs. 0.46 in the case of $k$-means).
  Interestingly, the correlation results are slightly better for $k$-means: the range of values for the pairwise distances ($x$-axis) is much larger, so it seems like the $k$-means guest profiles are better at capturing extremely divergent users, and the resulting ranking pairs are more divergent too.
  Table \ref{tab:ranking_example} presents an example of divergent rankings over review sentences for a given listing from two different guest profiles using the ABAE model.

  \begin{table*}[t!]
    \small
    \begin{center}
      \begin{tabular}{r|p{0.4\linewidth}p{0.4\linewidth}}
        {\bf Rank} & {\bf Guest 1} & {\bf Guest 2} \\
        \hline
        1 & Room is cozy and clean only the washroom feel a little bit old. &  Within walking distance to Feng Chia Night Market yet quiet enough when it comes time to rest.\\
        2 & Clean and comfortable room for the lone traveller or couples. &  Nice and clean place to stay, very near to Fengjia night market. \\
        3 &  The room is very good, as good as on the photos, and also clean. & Overall my TaiChung trip was good and really convenient place to stay at Nami's place. \\
        4 & Nice and clean place to stay, very near to Fengjia night market. & Ia a great place to stay, clean. \\
        5 & Within walking distance to Feng Chia Night Market yet quiet enough when it comes time to rest. & Near feng jia night market.
      \end{tabular}
    \end{center}
    \caption{\small From the experiment in \S\ref{sec:eval_ranking}, ranked review sentences for two different guest profiles for the same listing using the ABAE model. The first guest's profile focuses on the listing interior and cleanliness aspects, whereas the second guest is more interested in location.}
    \label{tab:ranking_example}
  \end{table*}

  \section{Conclusion}
  \label{sec:conclusion}
  In this work, we evaluated a recently proposed neural-based aspect extraction model in several settings.
  First, we used the inferred sentence-level aspects to select prototypical review sentences of a listing for a given aspect, and evaluated this aspect identification/extractive summarization task using human evaluators benchmarked against two baselines.
  Interestingly, the $k$-means baseline does quite well on frequently-occurring aspects.
  Second, the sentence-level aspects were also used to compute user profiles by grouping reviews that individual users have written.
  We showed that these embeddings are effective in reranking sentences, reviews, or listings in order to personalize this content to individual users.

  For future work, we wish to investigate alternative ways to aggregate and compute user profiles and compute distances between objects to rank and user profiles.
  We would also like to utilize human evaluators to judge the rankings produced in the review reranking experiments.

  \section*{Acknowledgments}
  We thank the authors of the ABAE paper \cite{He2017} for providing us with their code\footnote{\texttt{https://github.com/ruidan/} \newline \texttt{Unsupervised-Aspect-Extraction}}, which we used to benchmark our internal implementation.
  We are also grateful to our human evaluators.

  \bibliographystyle{acl_natbib}
  \bibliography{abae}

  \end{document}